\documentclass{article}

    \PassOptionsToPackage{numbers, compress}{natbib}
\usepackage[preprint]{neurips_2026}

\usepackage[utf8]{inputenc} 
\usepackage[T1]{fontenc}    
\usepackage{hyperref}       
\usepackage{url}            
\usepackage{booktabs}       
\usepackage{amsfonts}       
\usepackage{nicefrac}       
\usepackage{microtype}      
\usepackage{xcolor}         
\usepackage{placeins}
\usepackage{adjustbox} 
\renewcommand{\cite}[1]{\citep{#1}}
\usepackage[most]{tcolorbox}

\usepackage[most]{tcolorbox}
\usepackage{listings}
\usepackage{xcolor}

\newtcblisting{promptbox}[2][]{%
  enhanced,
  breakable,
  colback=white,
  colframe=black,
  arc=3mm,
  boxrule=0.8pt,
  width=\textwidth,
  title={\parbox{0.95\linewidth}{\textbf{#2}}},
  coltitle=white,
  colbacktitle=blue,
  fonttitle=\bfseries,
  left=6pt,
  right=6pt,
  top=6pt,
  bottom=6pt,
  before skip=8pt,
  after skip=8pt,
  listing only,
  listing options={
    basicstyle=\ttfamily\scriptsize,
    breaklines=true,
    breakatwhitespace=false,
    columns=fullflexible,
    keepspaces=true,
    showstringspaces=false,
    upquote=true,
    postbreak=\mbox{\textcolor{gray}{$\hookrightarrow$}\space}
  },
  #1
}

\title{Unveiling Fine-Grained Visual Traces: Evaluating Multimodal Interleaved Reasoning Chains in Multimodal STEM Tasks}

\usepackage{graphicx}  
\usepackage{booktabs}  
\usepackage{multirow}  
\usepackage{amsmath}   
\usepackage{amssymb}   

\usepackage{subfig}
\usepackage{wrapfig}

\usepackage{tcolorbox}
\tcbuselibrary{breakable}

\newcommand{\bench}{{\textsc{StepSTEM}}}

\author{
  Jing Jin$^{1,2}$\thanks{\ \ Equal contribution.} \quad
  Hao Liu$^3$\footnotemark[1] \quad
  Yan Bai$^2$\footnotemark[1] \quad
  Yihang Lou$^4$  \quad
  Zhenke Wang$^3$\quad
  Tianrun Yuan$^3$ \\
  \bfseries Juntong Chen$^3$ \quad
  Yongkang Zhu$^3$ \quad
  Fanhu Zeng$^5$ \quad
  Xuanyu Zhu$^{2,4}$ \quad
  Tao Feng$^{1}$ \quad
  Yige Xu$^6$\thanks{\ \ Corresponding author.} \\
  \\
  $^1$Tsinghua University  \quad \quad
  $^2$Meituan Inc  \quad \quad
  $^3$Central South University  \quad \quad
  $^4$Peking University \\
  $^5$University of Chinese Academy of Sciences  \quad \quad
  $^6$Nanyang Technological University \\
  \texttt{jingjin0007@gmail.com} \quad
  \texttt{liuhaokix@csu.edu.cn} \quad \texttt{yanbai02@meituan.com} \quad \\
  \texttt{xyzhu@stu.pku.edu.cn} \quad
  \texttt{yige002@ntu.edu.sg}
}

\begin{document}
\maketitle

\begin{abstract}
    Multimodal large language models (MLLMs) have shown promising reasoning abilities, yet evaluating their performance in specialized domains remains challenging. STEM reasoning is a particularly valuable testbed because it provides highly verifiable feedback, but existing benchmarks often permit unimodal shortcuts due to modality redundancy and focus mainly on final-answer accuracy, overlooking the reasoning process itself. To address this challenge, we introduce \textbf{\bench}: a graduate-level benchmark of 283 problems across mathematics, physics, chemistry, biology, and engineering for fine-grained evaluation of cross-modal reasoning in MLLMs. \textbf{\bench} is constructed through a rigorous curation pipeline that enforces strict complementarity between textual and visual inputs. We further propose a general step-level evaluation framework for both text-only chain-of-thought and interleaved image-text reasoning, using dynamic programming to align predicted reasoning steps with multiple reference solutions. Experiments across a wide range of models show that current MLLMs still rely heavily on textual reasoning, with even Gemini 3.1 Pro and Claude Opus 4.6 achieving only 38.29\% accuracy. These results highlight substantial headroom for genuine cross-modal STEM reasoning and position \textbf{\bench} as a benchmark for fine-grained evaluation of multimodal reasoning. Source code is available at \url{https://github.com/lll-hhh/STEPSTEM}.
\end{abstract}

\begin{figure*}[t]
    \centering
    \hspace*{-0.2cm}\includegraphics[width=\linewidth]{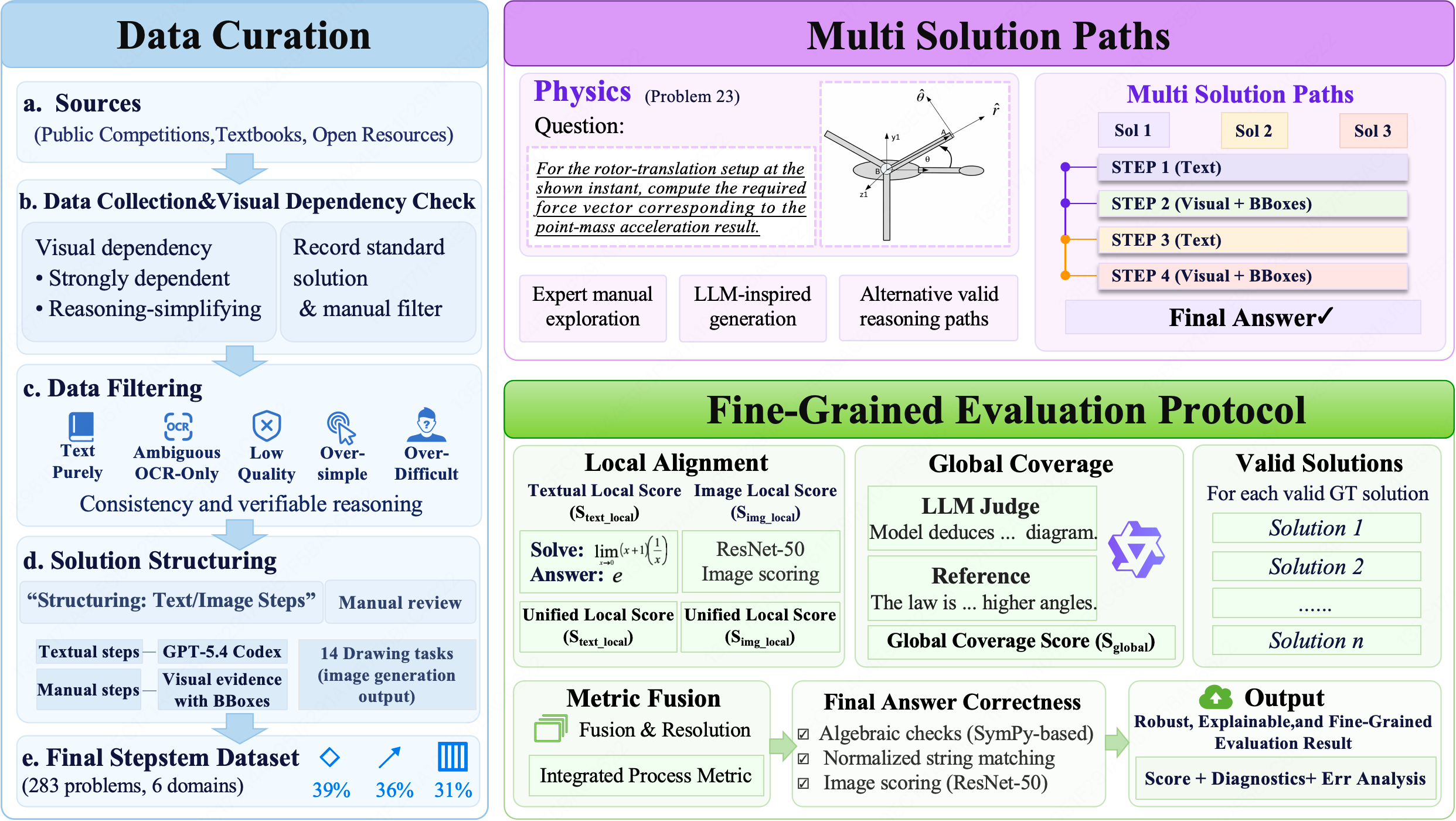}
    \caption{Overview of the {\bench} construction and evaluation workflow, including the data curation pipeline, the multi-solution construction pipeline, and the fine-grained evaluation protocol.}
    \label{fig:overview}
\end{figure*}

\section{Introduction}

Recent advances in Multimodal Large Language Models (MLLMs)~\cite{internvl3,qwen2026qwen35,bagel,janus,emu3,longcat} have greatly expanded the scope of multimedia understanding and generation. Beyond processing complex visual and textual inputs, MLLMs can now produce intermediate reasoning traces in textual, visual, or interleaved image-text forms, and use them to support final predictions~\cite{thinkwithimage,zebra,thinkwithgenerateimage}. This capability makes MLLMs a promising foundation for complex multimodal reasoning. Among such tasks, STEM reasoning (Science, Technology, Engineering, and Mathematical) is particularly important because final answers are often objectively verifiable, providing a natural testbed for analyzing both reasoning outcomes and reasoning processes.

Despite recent progress, multimodal STEM reasoning remains challenging. First, existing MLLMs often suffer from \textit{modality collapse}: they over-rely on textual cues while under-utilizing visual evidence, resulting in weak visual grounding and degraded performance on tasks that require genuine visual reasoning~\cite{DBLP:conf/emnlp/ZhaoSCZSCZ25,DBLP:conf/acl/WuSWH25,yao2025rethinking}. This issue is especially pronounced in STEM benchmarks, where detailed textual descriptions may allow models to bypass visual information. Second, current models still struggle to understand and generate structured STEM visuals. Scientific figures, diagrams, and plots are underrepresented in existing training data, and mainstream image-generation pipelines are largely optimized for naturalistic images rather than layouts with precise symbols, labels, and logical constraints~\cite{DBLP:conf/acl/0039WXWFK024,DBLP:conf/cvpr/Wei0CW0GSYG25}. Third, current evaluation and post-training practices mainly emphasize final-answer correctness, while providing limited insight into the quality and faithfulness of intermediate reasoning. Prior work has shown that chain-of-thought rationales may not faithfully reflect the underlying reasoning process, and that step-wise supervision can better support reliable reasoning than outcome-only signals~\cite{DBLP:conf/nips/TurpinMPB23,DBLP:conf/iclr/LightmanKBEBLLS24,zhang2025r1}. As a result, it remains difficult to determine whether a model truly performs grounded cross-modal reasoning or merely exploits superficial shortcuts.

To address these limitations, we introduce \textbf{\bench}, a multimodal STEM benchmark with fine-grained reasoning-step evaluation. {\bench} contains 283 graduate-level problems spanning engineering, physics, mathematics, chemistry, and biology. Each problem is carefully curated to require both textual information and visual evidence, thereby discouraging unimodal shortcuts and encouraging genuine cross-modal reasoning. As shown in Figure~\ref{fig:overview}, our data construction process enforces strict text-vision complementarity: solutions cannot be recovered from text alone or vision alone. Moreover, each problem is annotated with one or more ground-truth solution trajectories containing both textual and visual intermediate steps.

Meanwhile, we propose a \textit{step-level} evaluation framework for multimodal reasoning. Rather than evaluating only final-answer accuracy, our framework measures the similarity between predicted reasoning steps and annotated ground-truth steps, then aligns predicted trajectories with one or more reference trajectories using dynamic programming. The resulting score assesses not only whether a model obtains the correct answer, but also whether it follows a plausible, grounded, and interpretable multimodal reasoning path.

We evaluate a broad range of representative MLLMs on {\bench}. Experimental results show that existing models still struggle with graduate-level multimodal STEM reasoning, revealing substantial room for improvement. In addition, our step-level scores provide a more fine-grained view of model behavior than answer-only evaluation, exposing differences in reasoning quality, modality usage, and grounding ability that are obscured by final accuracy alone.

In summary, our contributions are three-fold:
{\bf (1) A multimodal STEM reasoning benchmark requiring cross-modal reasoning:} We introduce \textbf{\bench}, a benchmark of 283 carefully curated STEM problems that enforce strict text-vision complementarity to reduce unimodal shortcuts.
{\bf (2) A fine-grained framework for intermediate reasoning evaluation:}
We propose a \textbf{step-level evaluation framework} that systematically assesses intermediate multimodal reasoning steps beyond final-answer accuracy; and
{\bf (3) A comprehensive empirical analysis:}
We evaluate representative MLLMs on \textbf{\bench} and analyze their capabilities and limitations in cross-modal STEM reasoning. We further show that step-level scores strongly correlate with final-answer accuracy, highlighting the importance of evaluating reasoning processes in addition to final outcomes.

\section{Related Works}

\subsection{Multimodal Chain-of-Thought Reasoning}

Chain-of-Thought (CoT) prompting~\cite{DBLP:conf/nips/Wei0SBIXCLZ22} has achieved strong performance in text-based reasoning, but purely textual rationales provide limited grounding for problems that require spatial perception or diagram understanding. To address this limitation, recent work has explored interleaved multimodal CoT~\cite{rose2023visual,DBLP:journals/tmlr/0001Z00KS24}, where textual and visual inference are jointly used across intermediate reasoning steps.

Existing explicit multimodal CoT methods mainly follow two directions. The first is tool-augmented reasoning, which invokes external visual modules, drawing tools, or code interpreters to support intermediate reasoning~\cite{toolused1,toolused2,toolused3,toolused4,zhu-vtc,liyuying}. While effective in some settings, such pipelines are often sensitive to tool interfaces and error propagation, limiting their flexibility and robustness. The second direction studies intrinsic visual CoT within unified model architectures. For example, THINKMORPH~\cite{thinkmorph} investigates emergent reasoning abilities under multimodal interleaved training, while MathCanvas~\cite{mathcanvas} develops an intrinsic VCoT framework for mathematical reasoning. However, most existing text-to-image models are pre-trained on open-domain visual data and optimized for general-purpose image generation, rather than task-oriented visual generation for structured reasoning, which limits their effectiveness in downstream reasoning scenarios.

\subsection{Multimodal Reasoning Dataset}

High-quality multimodal reasoning data remains another key bottleneck. Early multimodal reasoning benchmarks mainly focus on visual understanding, where models are required to interpret visual inputs and produce textual answers. Representative datasets include MMMU~\cite{mmmu}, MathVista~\cite{mathvista}, ScienceQA~\cite{scienceqa}, and MathVerse~\cite{mathverse}. These benchmarks have substantially advanced multimodal model evaluation, but their solution processes are often text-centric and provide limited supervision for intermediate cross-modal reasoning. More recent unified benchmarks, such as Uni-MMMU~\cite{uni-mmmu}, RealUnify~\cite{realunify}, and others~\cite{quantifying,mmdeepresearch}, evaluate both perception and generation abilities. However, they mainly target foundational multimodal capabilities and typically lack the graduate-level depth and systematic step-level protocols needed to audit complex reasoning traces. Other works, like Zebra-CoT~\cite{zebra}, further explore reasoning with visual generation. Nevertheless, existing STEM benchmarks often suffer from \emph{modality redundancy}, where models can exploit textual shortcuts without genuinely relying on visual evidence. Many datasets also remain insufficiently challenging for evaluating deep, multi-step reasoning in state-of-the-art MLLMs.

\subsection{Fine-Grained Evaluation and Supervision of Reasoning Processes}

Most existing evaluation protocols emphasize final-answer accuracy, offering limited insight into the faithfulness, coherence, and grounding of intermediate reasoning steps. To address this issue, recent work has increasingly focused on process-level supervision. Process Reward Models, such as PRM800K~\cite{prm800k} and DREAM-PRM~\cite{dreamprm}, demonstrate the value of step-level feedback for improving reasoning reliability and reducing logical hallucinations. Meanwhile, the LLM-as-a-Judge paradigm, including  JudgeLM~\cite{judgelm} and Prometheus~\cite{prometheus}, has become a common framework for evaluating open-ended reasoning. Fine-grained evaluation has also begun to emerge in multimodal generation. Benchmarks such as T2I-CoReBench~\cite{T2I-CoReBench} and Science-T2I~\cite{sciencet2i} evaluate structured generated content beyond holistic image quality. Our work builds on this trend, but focuses on interleaved multimodal STEM reasoning. In this setting, evaluation must assess not only local step quality, but also whether the predicted reasoning trajectory aligns with a valid cross-modal solution path.

\begin{table*}[t]
\centering\tabcolsep 2pt\small
\caption{Detailed statistics of the {\bench}. The table presents the distribution of problems across various domains, the availability of multi-path solutions (the 2$^{\text{nd}}$ Solution and the 3$^{\text{rd}}$ Solution), and the average number of total, textual, visual steps, and bounding boxes per reference solution.}
\label{tab:dataset_statistics}
\begin{tabular}{@{}lccccccc@{}}
\toprule
\multirow{2}{*}{\textbf{Domain}} & \multirow{2}{*}{\textbf{\# Questions}} & \multicolumn{2}{c}{\textbf{Multi-Solution Samples}} & \multicolumn{4}{c}{\textbf{Average Steps per Solution}} \\ \cmidrule(lr){3-4} \cmidrule(l){5-8}
 &  & \textbf{Has 2$^{\text{nd}}$ Sol.} & \textbf{Has 3$^{\text{rd}}$ Sol.} & \textbf{Total Steps} & \textbf{Text Steps} & \textbf{Image Steps} & \textbf{BBoxes} \\ \midrule
Engineering & 111 & 34 & 14 & 7.6 & 4.4 & 3.2 & 5.1 \\
Physics     & 94  & 30 & 0  & 7.0 & 3.7 & 3.3 & 4.6 \\
Chemistry   & 25  & 3  & 0  & 6.2 & 3.2 & 3.0 & 6.1 \\
Mathematics & 30  & 1  & 0  & 6.9 & 3.6 & 3.3 & 4.9 \\
Other   & 23  & 4  & 0  & 7.0 & 3.6 & 3.4 & 5.7 \\ \midrule
\textbf{Overall} & \textbf{283} & \textbf{72} & \textbf{14} & \textbf{7.2} & \textbf{4.0} & \textbf{3.3} & \textbf{5.0} \\ \bottomrule
\end{tabular}
\end{table*}

\section{Methodology}
\label{sec:methodology}

\subsection{Overview}

\begin{wrapfigure}{r}{0.38\textwidth}
  \begin{center}
    \vspace{-15pt} %
    \includegraphics[width=0.38\textwidth]{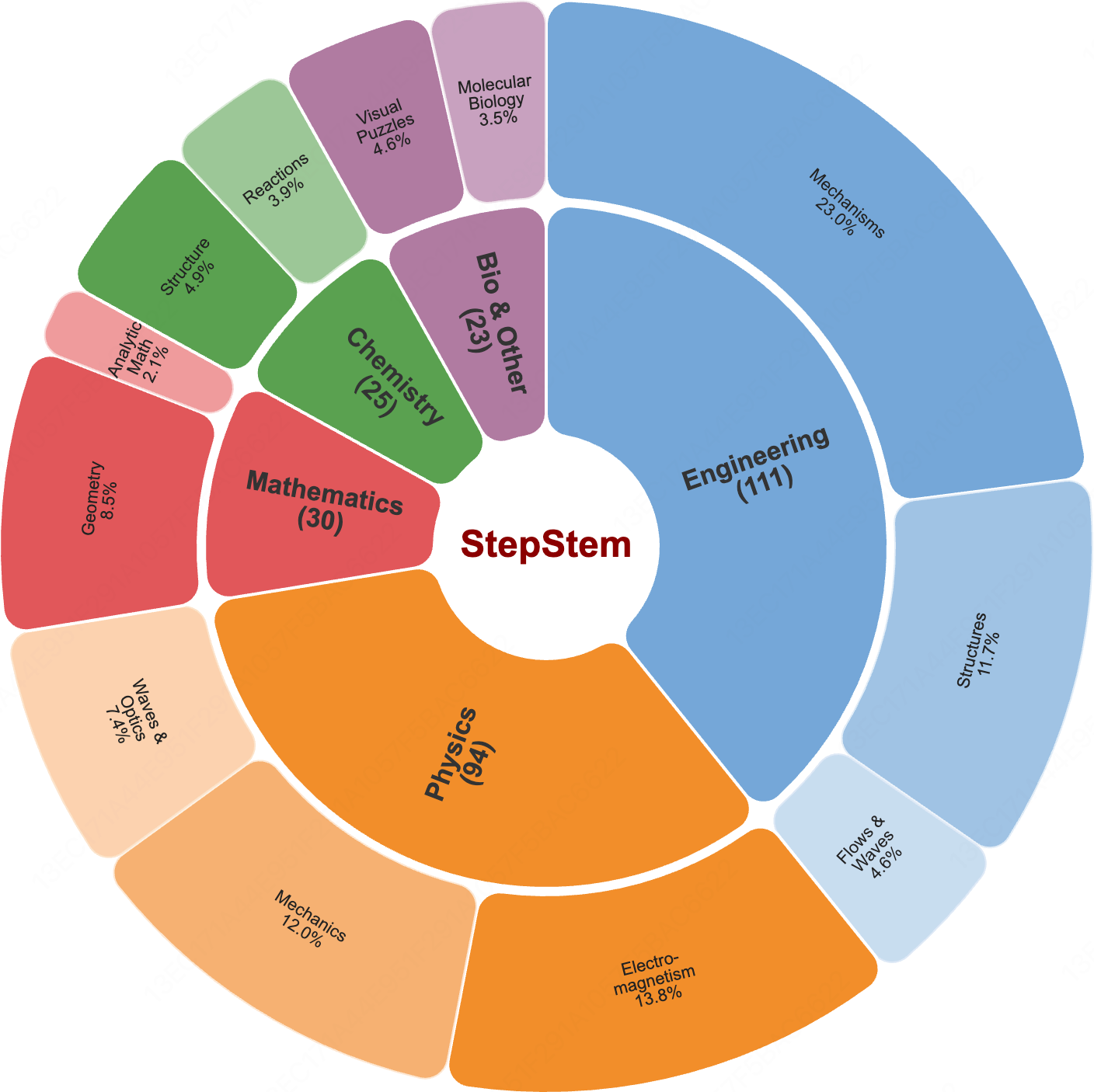}
    \vspace{-10pt} %
    \caption{Domain distributions of {\bench}.}
    \label{fig:bintu}
    \vspace{-35pt} %
  \end{center}
\end{wrapfigure}

Our {\bench} consists of two core components: a carefully curated multimodal STEM dataset (\S~\ref{sec:data_curation}) and a fine-grained evaluation protocol (\S~\ref{sec:eval_protocol}). For the dataset, we collect 283 graduate-level multimodal STEM reasoning problems and annotate each with one or more ground-truth reference solutions. The overall construction workflow is shown in Figure~\ref{fig:overview}, and the domain distribution of {\bench} is reported in Figure~\ref{fig:bintu}. For evaluation, our framework assesses both final-answer correctness and intermediate reasoning quality, including local step quality, global trajectory alignment, and their fusion. When multiple reference solutions are available, we compare the prediction against each reference independently and report the best score.

\subsection{Data Curation}
\label{sec:data_curation}

To support process-level evaluation, we curate problems with substantial visual dependency. Specifically, we include two types of samples: (1) \textit{strongly dependent} problems, where the solution cannot be reliably derived without the image; and (2) \textit{reasoning-simplifying} problems, where visual information substantially shortens the reasoning chain or reduces semantic ambiguity. The curation pipeline consists of data collection, data filtering, and solution structuring. This process was conducted by eight experts with interdisciplinary backgrounds, who converted raw multimodal problems into structured reasoning trajectories suitable for fine-grained evaluation. Detailed dataset statistics are in Table~\ref{tab:dataset_statistics}.

\paragraph{Data Collection}

Experts collected candidate problems from public competitions, graduate-level textbooks, and other open educational resources. For each candidate, we recorded the problem statement, reference image, ground-truth answer, and a standard solution trajectory containing key intermediate steps. We followed a task-attribute-first principle to assess visual dependency: a problem was retained only if solving it required geometric structures, spatial relations, or local visual details, or if the image substantially simplified the reasoning process. To broaden the evaluation setting, we also include 14 drawing tasks whose final output is an image. These tasks extend the benchmark from conventional ``image-text understanding to text generation'' toward the more challenging setting of ``image-text understanding to image generation.''

\paragraph{Data Filtering}
\label{sec:data_filtering}

We applied strict filtering criteria to ensure data quality and evaluation reliability. We removed samples where: (1) the image was purely decorative or redundant with the text; (2) the problem was ambiguous or mainly tested OCR ability; or (3) the image quality was too low to support stable annotation. We further applied a two-sided difficulty filter: problems that experts agreed required no explicit intermediate reasoning were discarded to avoid reducing the benchmark to basic visual recognition. Conversely, problems were also excluded if experts could not reach a stable consensus or formulate a verifiable step-level solution process.
The resulting domain distribution reflects different levels of intrinsic visual dependency across STEM fields. Engineering and Physics naturally contain more graduate-level problems that require spatial or geometric reasoning, whereas {\bf many candidate problems in Mathematics and Biology were removed because they could be solved through purely textual reasoning}. This filtering strategy helps reduce unimodal shortcuts and ensures that retained samples require meaningful use of visual evidence.

\paragraph{Solution Structuring}

We structure expert-verified solutions into interleaved image-text reasoning trajectories, using GPT-5.4 Codex only for standardized step decomposition and modality-tag assignment under fixed prompts, followed by manual review to ensure each step is independent and verifiable. For visual reasoning steps, experts annotate the relevant local evidence with one or more bounding boxes, enabling fine-grained supervision for step-level image alignment. For problems with multiple valid solution paths, we construct alternative references through expert exploration and LLM-assisted strategy proposal, retaining only those verified by experts as correct, logically consistent, and substantially distinct. As a result, multiple references in {\bench} capture genuinely different deductive paths rather than paraphrastic variants, supporting robust \textit{best-over-solutions} evaluation. More details for solution structuring are shown in Appendix~\ref{sec:appendix_solution_structuring}.

Through this pipeline, we construct {\bench}, a dataset of 283 problems spanning Engineering, Physics, Chemistry, Mathematics, Biology and Other domains.

\subsection{Evaluation Protocol}
\label{sec:eval_protocol}

Beyond final-answer accuracy, {\bench} evaluates the quality of intermediate multimodal reasoning traces. Each predicted trajectory is compared with one or more annotated reference solutions through two complementary signals. First, a local alignment score measures step-level correspondence: textual steps are matched using attention-rollout-based semantic similarity followed by monotonic dynamic programming, while visual steps are matched by comparing BBox-anchored reference features with sliding-window features from predicted images. A diversity penalty is applied to prevent a single generic generated image from satisfying multiple visual steps. Second, a global LLM judge assesses whether the full predicted trajectory covers the critical textual propositions and visual evidence required by the reference solution. The local and global scores are fused with a weighted RMS, and when multiple valid reference solutions are available, we report the maximum fused score over all references. Final-answer correctness is evaluated separately for textual and visual outputs. More technical details, including textual reasoning step evaluation, visual reasoning step evaluation, global logic coverage evaluation, metric fusion with final answer evaluation, and human evaluation are provided in Appendix~\ref{app:eval_protocol}.

\section{Experiments}
\label{sec:experiments}

\subsection{Baselines}
\label{sec:baselines}

To comprehensively evaluate the multimodal reasoning capabilities of current Large Vision-Language Models, we select a diverse set of representative baselines. As outlined in Table~\ref{tab:main_results}, these models are categorized into three distinct groups based on their architectural capabilities and accessibility:

\noindent \textbf{Open-Source Unified Models:} This group represents the frontier of open-weight models capable of both understanding and generating interleaved text and images. The complete list and detailed configurations are available in Appendix~\ref{sec:appendix_models}.

\noindent \textbf{Closed-source Unified Models:} This group represents the current state-of-the-art commercial systems capable of native multimodal generation. We evaluate Gemini 2.5 Flash Image \cite{google_gemini25_flash_image_2025} and a pipeline approach combining GPT-5.4 with GPT-image \cite{openai_gpt54_2026,openai_gpt_image_1}.

\noindent \textbf{Understanding-only Models:} To demonstrate the necessity of unified generation, we also evaluate powerful models restricted to textual output, including Claude Opus 4.6 \cite{anthropic_claude_opus46_2026}, Gemini 3.1 Pro \cite{google_gemini31_pro_2026}, and various scales of the Qwen family.

\begin{table*}[t]
  \centering
  \small
  \caption{Overall results on 269 problems with textual final answer, and 14 problems with image-based final answer. \textbf{Txt./Img. Acc.} evaluates the correctness of textual final answers or image-based final answers. \textbf{Local (T/I) }evaluates fine-grained alignment of intermediate textual and visual reasoning steps $S_{\text{local}}$. \textbf{Global (T/I)} assesses semantic coverage of reference key points and visual regions across the entire trace $S_{\text{global}}$. \textbf{Fused }is the integrated process score $S_{\text{fused}}$.}
  \label{tab:main_results}
  \resizebox{\textwidth}{!}{
  \begin{tabular}{l|ccc|ccc|ccc}
    \toprule
    Model & Txt. Acc. & Img. Acc. & Fused & Local-T & Local-I & Local & Global-T & Global-I & Global \\
    \midrule
    \multicolumn{10}{c}{\textit{Unified Models}} \\
    \midrule
    Gemini 2.5 Flash Image & 15.61 & {\bf 50.80} & 67.05 & 58.62 & {\bf 64.10} & {\bf 62.35} & 65.57 & 78.56 & 71.40 \\
    GPT-5.4 + GPT-Image & 33.21 & 39.40 & {\bf 71.52} & 57.72 & 61.83 & 60.20 & 74.46 & {\bf 88.21} & {\bf 80.94} \\
    \midrule
    \multicolumn{10}{c}{\textit{Interleaved Generation Models}} \\
    \midrule
    Bagel & 4.83 & 0.00 & 47.38 & 30.64 & 54.15 & 44.63 & 50.34 & 48.70 & 49.76 \\
    Janus-Pro & 0.37 & 0.00 & 40.99 & 24.20 & 53.20 & 49.52 & 38.72 & 19.16 & 30.11 \\
    OmniGen2 & 3.35 & 0.00 & 45.04 & 23.58 & 59.05 & 42.30 & 45.97 & 45.91 & 47.76 \\
    Showo2 & 1.86 & 0.00 & 42.17 & 35.39 & 54.13 & 47.44 & 41.58 & 28.33 & 36.13 \\
    \midrule
    \multicolumn{10}{c}{\textit{Understanding-only Models}} \\
    \midrule
    Gemini 3.1 Pro & {\bf 38.29} & 0.00 & 38.13 & {\bf 66.13} & 0.00 & 36.65 & 71.49 & 0.00 & 39.53 \\
    Claude Opus 4.6 & {\bf 38.29} & 0.00 & 37.87 & 61.38 & 0.00 & 34.15 & {\bf 74.98} & 0.00 & 41.25 \\
    Qwen3.5-122B & 34.94 & 0.00 & 36.55 & 64.72 & 0.00 & 35.45 & 68.78 & 0.00 & 37.43 \\
    Qwen3.5-Plus & 29.00 & 0.00 & 35.01 & 56.62 & 0.00 & 31.05 & 70.46 & 0.00 & 38.46 \\
    Qwen3-VL-235B & 23.42 & 0.00 & 33.18 & 64.65 & 0.00 & 34.26 & 58.61 & 0.00 & 32.05\\
    \bottomrule
  \end{tabular}
  }
\end{table*}

\begin{table*}[t]
  \centering
  \small
  \caption{Overall results under the split of correctness. For example, ``81.43 / 64.73'' indicates the averaged fused score for correct testing examples is 81.43, while the fused score for the incorrect testing examples is 64.73.}
  \label{tab:correct_wrong_split}
  \resizebox{\textwidth}{!}{
  \setlength{\tabcolsep}{1.2pt}
  \begin{tabular}{l|c|c|c|c|c|c|c}
    \toprule
    Model & Fused & Local-T & Local-I & Local & Global-T & Global-I & Global \\
    \midrule
    \multicolumn{8}{c}{\textit{Unified Models}} \\
    \midrule
    Gemini 2.5 Flash  & 81.43 / 64.73 & 60.46 / 58.29 & 68.74 / 63.79 & 65.21 / 61.59 & 88.29 / 62.14 & 100.00 / 75.70 & 95.21 / 68.11 \\
    GPT-5.4 + Image & 78.41 / 68.81 & 63.34 / 54.79 & 68.28 / 58.60 & 62.78 / 58.92 & 87.21 / 68.84 & 95.69 / 87.69 & 91.13 / 77.58 \\
    \midrule
    \multicolumn{8}{c}{\textit{Interleaved Generation Models}} \\
    \midrule
    Bagel & 65.84 / 47.46 & 32.35 / 30.98 & 67.25 / 54.26 & 51.80 / 45.00 & 76.92 / 49.79 & 80.13 / 49.38 & 77.29 / 49.94 \\
    Janus-Pro & 78.88 / 40.82 & 49.58 / 24.69 & 53.94 / 53.09 & 50.37 / 49.52 & 100.00 / 38.18 & 100.00 / 18.18 & 100.00 / 29.42 \\
    OmniGen2 & 52.17 / 44.80 & 26.32 / 22.85 & 58.77 / 59.34 & 45.17 / 43.45 & 63.89 / 43.50 & 50.00 / 45.54 & 58.07 / 46.04 \\
    Showo2 & 45.15 / 43.07 & 49.95 / 35.87 & 48.32 / 55.17 & 51.00 / 48.32 & 46.67 / 40.23 & 20.00 / 29.40 & 38.33 / 37.15 \\
    \midrule
    \multicolumn{8}{c}{\textit{Understanding-only Models}} \\
    \midrule
    Gemini 3.1 Pro & 42.86 / 35.33 & 71.26 / 63.54 & 0.00 / 0.00 & 38.42 / 35.71 & 87.27 / 61.74 & 0.00 / 0.00 & 46.97 / 34.37 \\
    Claude Opus 4.6 & 41.25 / 35.98 & 67.43 / 59.15 & 0.00 / 0.00 & 36.02 / 33.49 & 86.80 / 68.53 & 0.00 / 0.00 & 46.44 / 38.39 \\
    Qwen3.5-122B & 42.39 / 33.63 & 70.21 / 62.29 & 0.00 / 0.00 & 36.85 / 34.78 & 89.88 / 57.89 & 0.00 / 0.00 & 47.33 / 32.34 \\
    Qwen3.5-Plus & 41.72 / 31.00 & 68.74 / 51.91 & 0.00 / 0.00 & 36.24 / 28.98 & 88.27 / 60.94 & 0.00 / 0.00 & 46.45 / 34.13 \\
    Qwen3-VL-235B & 41.43 / 31.02 & 70.60 / 62.72 & 0.00 / 0.00 & 37.53 / 33.26 & 86.08 / 50.60 & 0.00 / 0.00 & 45.07 / 28.17 \\
    \bottomrule
  \end{tabular}
  }
\end{table*}

\subsection{Main Results}
\label{sec:main_results}

Table~\ref{tab:main_results} reports overall performance on 269 problems with textual final answers and 14 problems with image-based final answers. We group the evaluated MLLMs into three categories according to their reasoning paradigms: unified models, interleaved generation models, and understanding-only models.

\paragraph{The benchmark remains highly challenging.}
Overall results show that {\bench} is difficult even for strong contemporary MLLMs. On the 269 text-answer problems, the best accuracy is only 38.29\%, achieved by Gemini 3.1 Pro and Claude Opus 4.6. On the 14 image-answer problems, the highest score is 50.80\%, achieved by Gemini 2.5 Flash Image. These results indicate that current MLLMs remain far from reliably solving graduate-level multimodal STEM reasoning tasks.
\paragraph{Vision-text-to-text models exhibit a fundamental limitation.}
Understanding-only models, such as Claude Opus 4.6 and Gemini 3.1 Pro, achieve relatively strong scores on text-related process metrics, particularly on Local-T and Global-T. However, all such models obtain zero scores on Img. Acc., Local-I, and Global-I. This pattern reveals a fundamental limitation of vision-text-to-text models: while they can process multimodal inputs and produce textual reasoning traces, they cannot handle tasks that require explicit visual generation as part of the reasoning process. Therefore, evaluating only the final answer is insufficient for complex multimodal STEM reasoning, since it fails to capture whether a model can produce and leverage image-based intermediate derivations.
\paragraph{Unified models achieve the strongest multimodal process scores but remain weak at final problem solving.}
Unified models outperform other categories on fused and global process metrics. GPT-5.4 + GPT-Image achieves the best Fused score of 71.52 and the highest Global score of 80.94, while Gemini 2.5 Flash Image obtains the best Img. Acc. score of 50.80\% and a strong Local-I score of 64.10. However, their final-answer accuracy remains limited, particularly on the text-answer subset: GPT-5.4 + GPT-Image reaches 33.21\%, while Gemini 2.5 Flash Image reaches only 15.61\%. These results suggest that current unified models can generate plausible intermediate visual structures, but still struggle to effectively exploit them for solving challenging STEM reasoning problems.
\paragraph{Interleaved generation alone is insufficient.}
Interleaved generation models obtain non-trivial image-oriented process scores, indicating some ability to produce intermediate visual content. For example, OmniGen2 achieves 59.05 on Local-I, and Bagel reaches 48.70 on Global-I. However, both models fail to produce correct image-based final answers, yielding 0\% accuracy on the image-answer subset. This suggests that simply inserting image generation into the reasoning chain is not enough. Effective multimodal STEM reasoning requires models to generate relevant visual evidence and integrate it coherently into the solution process.

\begin{figure*}[t]
    \centering
    \includegraphics[width=0.82\linewidth]{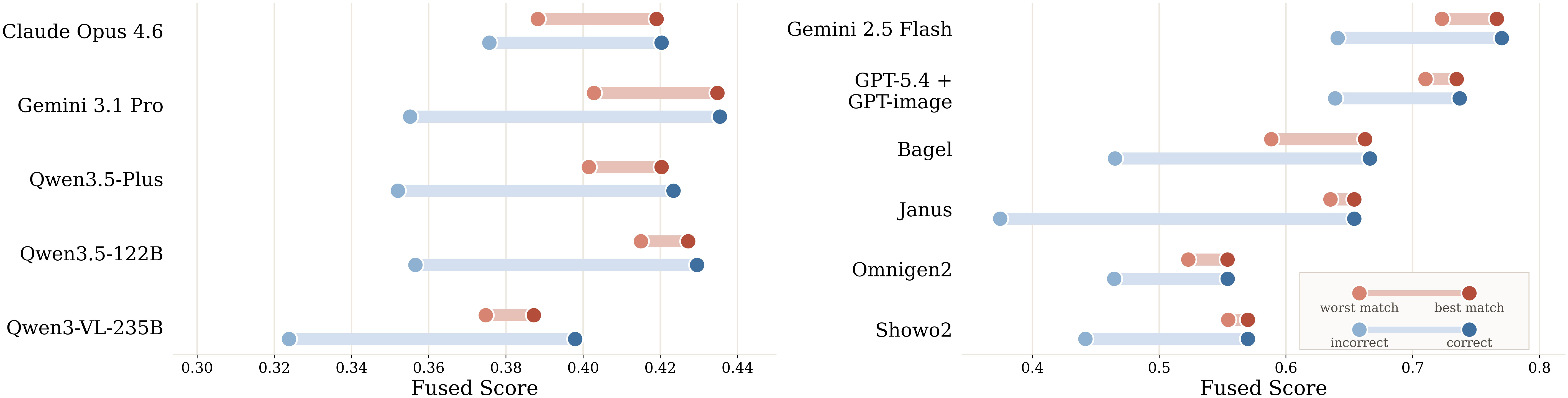}
    \caption{Comparison of the fused process score $S_{\text{fused}}$ across different model architectures. The red lines represent the gap between the best-matched and worst-matched ground truth solutions. The blue lines indicate the score difference between trajectories with correct and incorrect final answers.\label{fig:solution_gap_models_fused}}
\end{figure*}

\subsection{Analysis and Case Study}
\label{sec:analysis}

\subsubsection{Analysis of Step-Level Evaluation Scores}

\paragraph{Step-level process scores are strongly associated with final-answer correctness.}
As shown by the blue intervals in Figure~\ref{fig:solution_gap_models_fused}, trajectories that lead to correct final answers consistently receive higher fused process scores than those ending with incorrect answers. This pattern is further confirmed by the correctness split in Table~\ref{tab:correct_wrong_split}. For example, the fused score increases from 64.73 to 81.43 for Gemini~2.5~Flash, from 68.81 to 78.41 for GPT-5.4 + GPT-Image, from 47.46 to 65.84 for Bagel, and from 35.33 to 42.86 for Gemini~3.1~Pro when moving from incorrect to correct cases. The consistency of this trend across unified models, interleaved generation models, and understanding-only models suggests that the proposed process score captures meaningful variation in reasoning quality rather than incidental overlap with the reference solutions.

\paragraph{Higher-performing models tend to obtain higher process scores.}
Beyond the correctness split, we also observe that models with stronger multimodal reasoning ability generally achieve higher absolute process scores. In Table~\ref{tab:main_results}, the two unified models obtain the highest fused scores overall, indicating that stronger reasoning performance is accompanied by better intermediate trajectories. A similar trend is visible in Figure~\ref{fig:solution_gap_models_fused}, where better-performing models are generally located farther to the right on the fused-score axis. This observation provides further evidence for the validity of our step-level evaluation framework: stronger models do not merely produce more correct final answers, but also exhibit higher-quality reasoning processes.

\begin{figure*}[t]
    \centering
    \includegraphics[width=\linewidth]{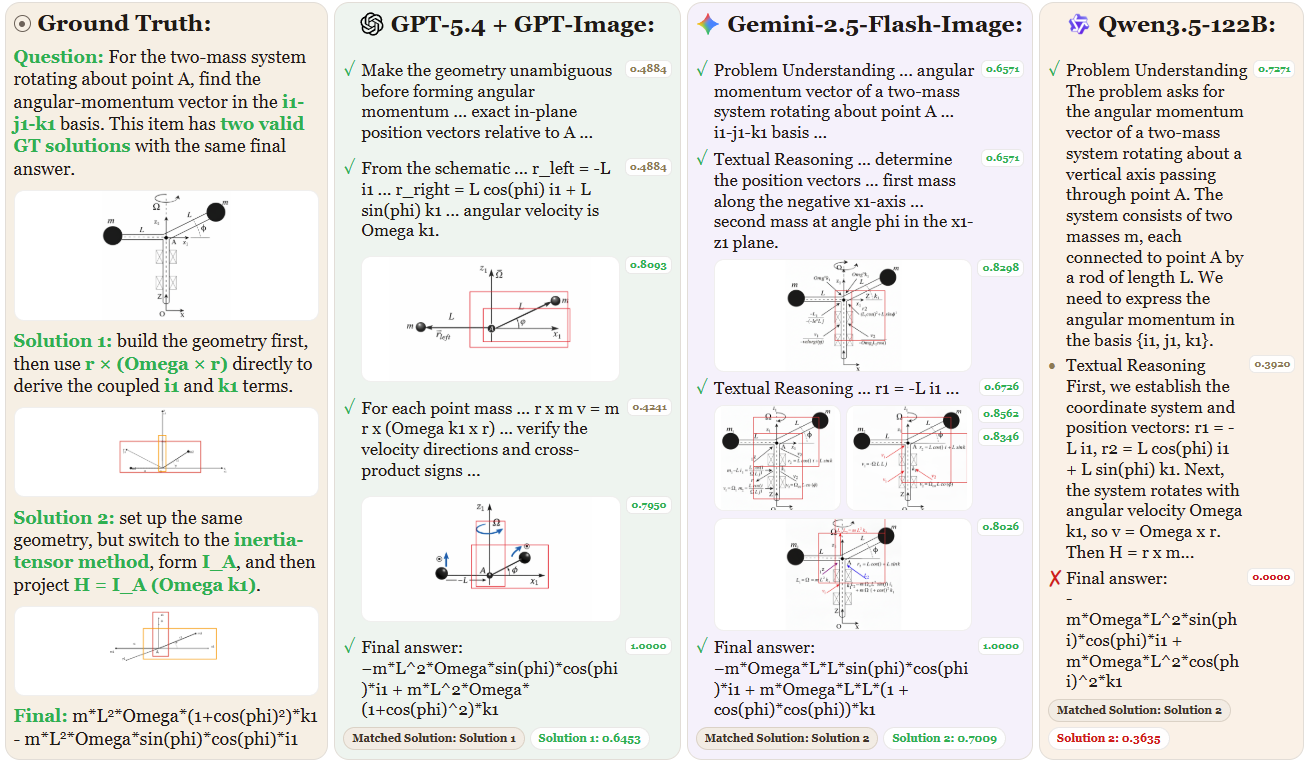}
    \caption{A qualitative case study demonstrating the model's predicted reasoning trajectory versus the ground-truth solutions. Our protocol successfully leverages the multi-solution mechanism and penalizes text-only derivations that lack essential geometric evidence.}
    \label{fig:case_study_eng_problem1}
\end{figure*}

\paragraph{The correct--incorrect gap is itself informative.}
The magnitude of the blue gap in Figure~\ref{fig:solution_gap_models_fused} also reveals differences in model behavior. Models such as Bagel and Janus-Pro show relatively large separations between correct and incorrect cases, indicating that successful predictions are supported by substantially better reasoning trajectories. In contrast, weaker models such as Showo2 and OmniGen2 exhibit much smaller gaps, suggesting that their intermediate trajectories remain of limited quality regardless of whether the final answer happens to be correct. This result highlights an important advantage of step-level evaluation: it can distinguish between models that occasionally reach the correct answer and models that more consistently follow a plausible reasoning process.

Overall, these results demonstrate the necessity of evaluating multimodal reasoning at the step level. Final-answer accuracy only indicates whether a model reaches the correct endpoint, but does not reveal whether the model follows a coherent cross-modal solution path. In contrast, our step-level evaluation scores provide a finer-grained diagnostic signal for assessing the quality of intermediate reasoning trajectories. This is particularly important for challenging multimodal STEM problems, where success depends not only on the final prediction, but also on whether the model can sustain a logically consistent reasoning chain across textual and visual steps.

\subsubsection{Analysis of Multiple Ground-Truth Solutions}

For problems with multiple valid solution strategies, we compute the process score against each ground-truth solution and examine the resulting distribution. As shown by the red intervals in Figure~\ref{fig:solution_gap_models_fused}, the same model trajectory can receive substantially different scores under different references. This variation does not necessarily indicate unstable model behavior; rather, it reflects the fact that complex multimodal STEM problems often admit multiple legitimate reasoning paths. A model may therefore follow a correct and coherent solution strategy while still obtaining a low score against one particular annotated reference.

This observation highlights a limitation of single-reference process evaluation. When only one ground-truth solution is available, the metric may unfairly penalize trajectories that are valid but structurally different from the chosen annotation. To mitigate this issue, we provide multiple ground-truth solutions whenever alternative solving strategies exist. With this design, a model can obtain a high process score as long as its reasoning is close to at least one valid reference path. This improves both the stability and the fairness of the evaluation framework by reducing spurious low scores caused by annotation incompleteness.

Moreover, the gap between the best-matched and worst-matched references is itself informative. A larger gap suggests greater diversity in valid solution trajectories and indicates that multi-reference evaluation is particularly important for such problems. Overall, these results show that multiple ground-truth solutions are essential for reliable step-level evaluation of multimodal reasoning.

\subsubsection{Case Study}

alid solution strategies, rather than overfitting to a single annotated chain of thought.

Figure~\ref{fig:case_study_eng_problem1} shows a representative example involving a two-mass system rotating about point~$A$, where the goal is to derive the angular-momentum vector in the $\mathbf{i}_1$-$\mathbf{j}_1$-$\mathbf{k}_1$ basis. The problem has two valid reference trajectories with the same final answer. Solution~1 directly computes $\mathbf{r}\times(\boldsymbol{\Omega}\times\mathbf{r})$ for each mass, while Solution~2 reformulates the problem using the inertia tensor and derives $\mathbf{H}$ through $\mathbf{I}_A(\Omega\mathbf{k}_1)$. This example illustrates a central challenge in step-level evaluation: correct solutions may follow substantially different intermediate structures. Our framework successfully accommodates such diversity. GPT-5.4 + GPT-Image is aligned with Solution~1, obtaining a matched-solution score of 0.6453 and a perfect final-answer score. In contrast, Gemini~3.1~Pro is aligned with Solution~2, achieving a matched-solution score of 0.7009 while also producing the correct final answer. Thus, models are not forced to mimic the same annotated reasoning chain; instead, they are rewarded when their trajectory is semantically close to any valid reference solution. This demonstrates that our multi-reference evaluation avoids over-penalizing alternative but correct reasoning strategies. The case also highlights the diagnostic value of step-level evaluation for incorrect predictions. Qwen3.5-122B obtains a relatively high problem-understanding score of 0.7271, indicating that it identifies the relevant physical setting and quantities. It also partially aligns with Solution~2 during intermediate reasoning. However, the model fails to recover the complete $\mathbf{k}_1$ component in the final angular-momentum vector, leading to an incorrect final expression and a zero final-answer score. Consequently, its matched-solution score drops to 0.3635. Such a failure pattern would be obscured by final-answer accuracy alone.

Overall, this case study demonstrates two advantages of \bench{}. First, multi-reference annotation enables fair process evaluation by recognizing distinct but valid reasoning paths. Second, step-level scoring provides a more informative diagnosis of model behavior, revealing whether errors arise from problem misunderstanding, flawed intermediate derivation, or final-step composition. These observations further support the need for fine-grained, multi-reference evaluation in multimodal STEM reasoning.

\begin{wrapfigure}{r}{0.45\textwidth}
  \begin{center}
    \vspace{-15pt} 
    \includegraphics[width=0.38\textwidth]{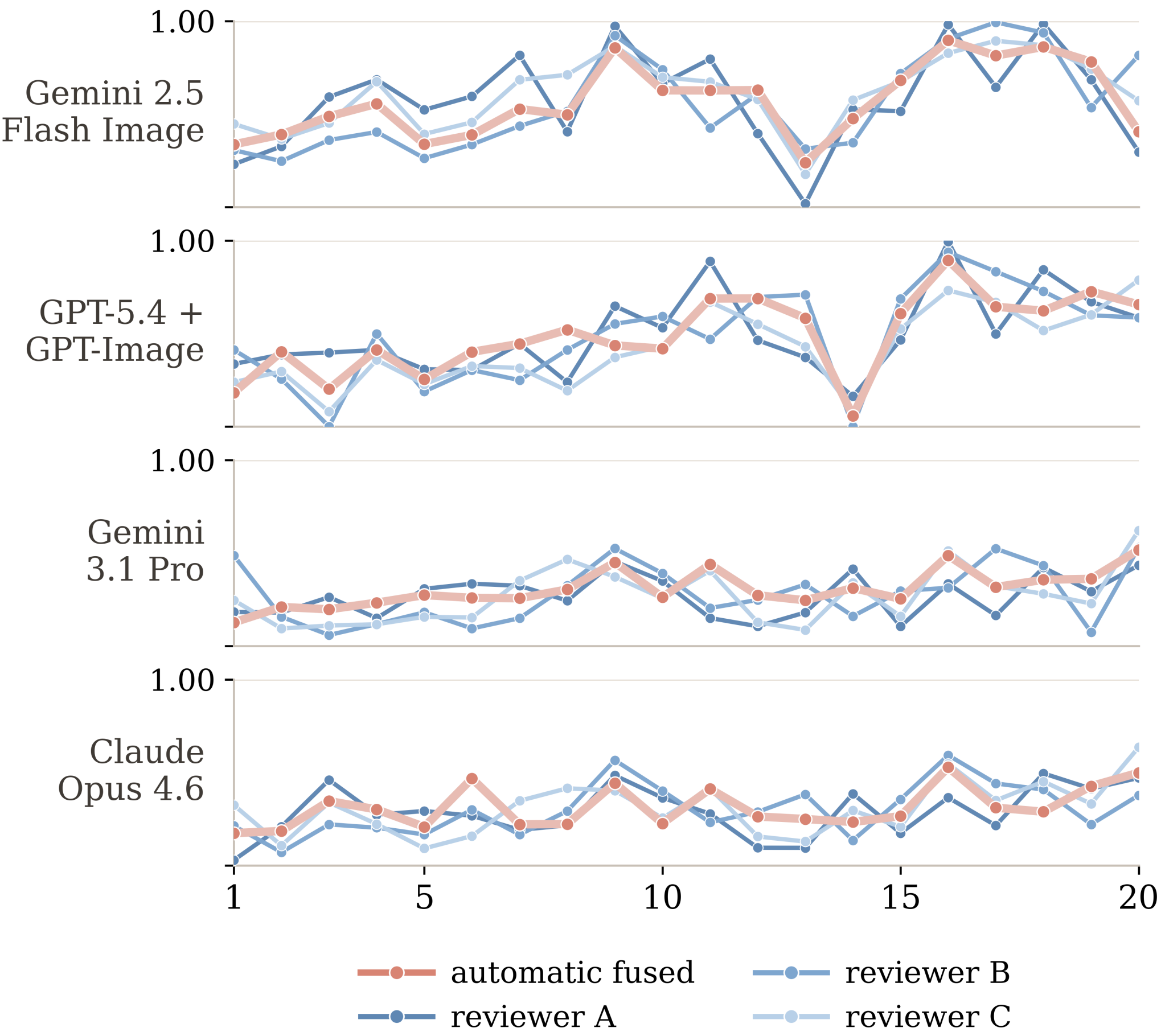}
    \vspace{-4pt} 
    \caption{Results for human evaluation.}
    \label{fig:human_evaluation}
    \vspace{-4pt} 
  \end{center}
\end{wrapfigure}

\subsection{Human Evaluation of Reasoning Paths}
\label{sec:human_eval}

Figure~\ref{fig:human_evaluation} compares the automatic fused score with the scores from the three human experts. The automatic scores show good qualitative agreement with human annotations. Across all four models, the automatic curve generally follows the same rises and drops as the three human curves, indicating that the proposed protocol captures how far a model can proceed along a valid reasoning path rather than merely measuring superficial overlap with the reference solution. High-scoring trajectories under human evaluation are usually assigned high automatic scores, while trajectories that fail in early reasoning steps receive consistently lower automatic scores. These results indicate that our automatic step-level evaluation is broadly aligned with expert judgement, while also highlighting the importance of multi-solution references and future improvements for handling compressed or semantically equivalent reasoning paths. More results for human evaluations are provided in Appendix~\ref{sec:appendix_human_eval}.

\section{Conclusions}

We introduced {\bench}, a benchmark and evaluation framework for interleaved multimodal STEM reasoning. By combining expert-curated problems, step-level textual and visual annotations, and multi-solution trajectory matching, {\bench} enables evaluation of not only whether a model reaches the correct final answer, but also how it reasons across modalities. Experiments show that current state-of-the-art MLLMs still exhibit a substantial gap between answer correctness and reasoning trajectories quality: to-text MLLMs cannot generate vision intermediate steps, whereas unified generative models better expose multimodal reasoning behavior but remain far from reliable. We hope {\bench} will support future research on multimodal process supervision, reasoning-oriented data construction, and more faithful evaluation of cross-modal reasoning systems.

\bibliography{custom}
\bibliographystyle{plainnat}

\appendix
\section*{Appendix}

\section{Methodology Details}
\subsection{Solution Structuring}
\label{sec:appendix_solution_structuring}

After filtering (\S~\ref{sec:data_filtering}), expert-verified solutions were converted into interleaved image-text reasoning trajectories. To improve consistency and efficiency, we used GPT-5.4 Codex with fixed prompts for initial step decomposition and modality-tag assignment. The model was used only to format expert-provided solutions, not to generate new answers. All structured outputs were then manually reviewed and corrected to ensure that each step forms an independent and verifiable reasoning unit.

For visual reasoning steps, experts annotated the corresponding local visual evidence with bounding boxes. When a step depended on multiple visual regions, multiple bounding boxes were allowed. These annotations identify the specific visual evidence used at each step and provide fine-grained supervision for step-level image alignment evaluation.

For problems admitting multiple valid solution paths, we further constructed alternative references through expert exploration and LLM-assisted proposal. Experts first identified diverse solution strategies based on their domain knowledge. In parallel, we queried advanced multimodal models such as Gemini 3.1 Pro and Claude Opus 4.6 to discover potential strategies that differed from the primary reference solution. All alternative solutions, whether proposed by experts or models, were retained only after expert verification confirmed that they were correct, logically consistent, and substantially distinct from existing references. Thus, multiple references in {\bench} represent genuinely different deductive paths rather than surface-level paraphrases, supporting the subsequent \textit{best-over-solutions} evaluation.

\subsection{Evaluation Protocol}
\label{app:eval_protocol}

We evaluate multimodal reasoning traces at two complementary levels:
\emph{local step alignment}, which measures whether individual predicted steps correspond to annotated reference steps, and \emph{global logic coverage}, which measures whether the overall trajectory covers the essential propositions and visual evidence required by the reference solution. This design allows the evaluation to reward explicit step-by-step agreement while remaining robust to valid alternative derivations that may not follow the reference trajectory verbatim.

For each problem, let $\mathcal{G} = \{g_1,\ldots,g_m\}$ denote a ground-truth reference trajectory and let $\mathcal{P} = \{p_1,\ldots,p_n\}$ denote the predicted trajectory. Each step is assigned a modality tag. We use $\mathcal{G}^T$ and $\mathcal{G}^I$ to denote the textual and visual reference steps, respectively, and $\mathcal{P}^T$ and $\mathcal{P}^I$ to denote the corresponding predicted steps. Textual reference steps are further annotated with key semantic points, while visual reference steps are annotated with one or more bounding boxes (BBoxes) indicating the local regions that provide necessary visual evidence.

\subsubsection{Textual Reasoning Step Evaluation}
\label{app:eval_text}

For textual reasoning, we prioritize semantic process coverage over surface-level string similarity. Given a reference textual step $g_i \in \mathcal{G}^T$ and a predicted textual step $p_j \in \mathcal{P}^T$, we compute a pairwise alignment score using attention rollout~\citep{DBLP:conf/acl/AbnarZ20} from Qwen3.5-9B~\citep{qwen2026qwen35}. Let $W^{(l,h)}$ denote the attention matrix of the $h$-th head in the $l$-th transformer layer. We first average attention over heads, add the residual connection, and normalize the resulting matrix:
\begin{equation}
\tilde{A}^{(l)}
=
\operatorname{RowNorm}
\left(
\frac{1}{H}\sum_{h=1}^{H} W^{(l,h)} + I
\right),
\end{equation}
where $H$ is the number of attention heads and $I$ is the identity matrix. The global attention rollout matrix is then obtained by multiplying the layer-wise matrices:
\begin{equation}
\bar{A}
=
\prod_{l=1}^{L} \tilde{A}^{(l)},
\end{equation}
where $L$ is the number of transformer layers.

To obtain a fine-grained similarity score between $g_i$ and $p_j$, we aggregate token-level dependencies from the rollout matrix. For each reference token $u \in g_i$, we identify the predicted token $v \in p_j$ with the maximum rollout weight and average these maxima across all reference tokens:
\begin{equation}
M^T[i,j]
=
\frac{1}{|g_i|}
\sum_{u \in g_i}
\max_{v \in p_j}
\bar{A}_{u,v}.
\end{equation}
This yields a textual step-similarity matrix $M^T$.

We then perform monotonic dynamic programming over $M^T$ to find the optimal alignment path $\pi^{T,*}$. The monotonicity constraint permits skipped steps, allowing models to merge or omit trivial intermediate derivations, while preventing crossed alignments that would violate the temporal order of the reasoning chain. The textual local score is defined as the average similarity over the matched pairs in the optimal path:
\begin{equation}
S_{\mathrm{local}}^{\mathrm{text}}
=
\frac{1}{|\pi^{T,*}|}
\sum_{(i,j)\in \pi^{T,*}}
M^T[i,j].
\end{equation}
If no valid textual prediction is available, we set $S_{\mathrm{local}}^{\mathrm{text}}=0$.

\subsubsection{Visual Reasoning Step Evaluation}
\label{app:eval_image}

For visual reasoning steps, such as intermediate diagrams, geometric constructions, plots, or state-transition figures, we evaluate whether the predicted images contain the visual evidence specified by the reference solution. Each reference visual step $g_i \in \mathcal{G}^I$ is associated with expert-annotated BBoxes. We encode the annotated region or regions using a visual encoder $\phi(\cdot)$ and use the resulting representation as the reference visual feature. When multiple BBoxes are annotated for the same step, their features are pooled to form the step-level representation.

Predicted visual steps are not assumed to contain annotations. Therefore, for each predicted image step $p_j \in \mathcal{P}^I$, we construct a set of sliding-window patches $W(p_j)$. The pairwise visual similarity between $g_i$ and $p_j$ is defined as the maximum cosine similarity between the reference visual feature and any patch feature extracted from the predicted image:
\begin{equation}
M^I[i,j]
=
\max_{w \in W(p_j)}
\frac{
\phi(g_i) \cdot \phi(w)
}{
\|\phi(g_i)\| \, \|\phi(w)\|
}.
\end{equation}
This produces the visual step-similarity matrix $M^I$.

A direct maximum-over-windows formulation may overestimate performance when a model repeatedly generates a single generic visual artifact that partially matches many reference steps. To discourage this degenerate behavior, we introduce a diversity-aware penalty. Let
\begin{equation}
C_{\mathrm{img}}
=
\frac{1}{|\mathcal{G}^I|}
\sum_{i=1}^{|\mathcal{G}^I|}
\max_{j} M^I[i,j]
\end{equation}
denote the maximum visual coverage over all reference visual steps. We further define the image diversity ratio as
\begin{equation}
D_{\mathrm{img}}
=
\frac{
\left|
\left\{
\arg\max_{j} M^I[i,j]
\;:\;
i=1,\ldots,|\mathcal{G}^I|
\right\}
\right|
}{
|\mathcal{G}^I|
},
\end{equation}
which measures how many distinct predicted visual steps are selected as best matches for the reference visual steps. The final visual local score is computed as
\begin{equation}
S_{\mathrm{local}}^{\mathrm{img}}
=
C_{\mathrm{img}}
\left(1-\lambda(1-D_{\mathrm{img}})\right),
\end{equation}
where $\lambda$ controls the strength of the diversity penalty. If a problem requires visual reasoning but the model produces no visual step, we set $S_{\mathrm{local}}^{\mathrm{img}}=0$.

The unified local process score combines textual and visual local scores according to the modality composition of the reference trajectory:
\begin{equation}
S_{\mathrm{local}}
=
\rho_T S_{\mathrm{local}}^{\mathrm{text}}
+
\rho_I S_{\mathrm{local}}^{\mathrm{img}},
\end{equation}
where
\begin{equation}
\rho_T =
\frac{|\mathcal{G}^T|}
{|\mathcal{G}^T|+|\mathcal{G}^I|},
\qquad
\rho_I =
\frac{|\mathcal{G}^I|}
{|\mathcal{G}^T|+|\mathcal{G}^I|}.
\end{equation}

\subsubsection{Global Logic Coverage Evaluation}
\label{app:eval_global}

Local alignment evaluates explicit step-level correspondence, but valid reasoning trajectories may differ substantially in structure from the reference solution. To account for such diversity, we additionally introduce an LLM-based global judge that evaluates whole-trajectory coverage.

For each reference solution, we collect a set of critical reference points $\mathcal{R}$, including textual key points and visual evidence specified by BBoxes. The full predicted trajectory is provided to the judge together with these reference points. The judge determines whether each reference point is semantically entailed, mathematically derived, or visually represented in the predicted trajectory. For each $r \in \mathcal{R}$, the judge outputs a binary coverage label:
\begin{equation}
c(r,\mathcal{P}) \in \{0,1\},
\end{equation}
where $1$ indicates that the reference point is covered and $0$ indicates that it is missing. The global coverage score is then computed as
\begin{equation}
S_{\mathrm{global}}
=
\frac{1}{|\mathcal{R}|}
\sum_{r \in \mathcal{R}}
c(r,\mathcal{P}).
\end{equation}
This coverage-oriented judgment rewards trajectories that contain the necessary reasoning content, even when the local step order or granularity differs from the annotated reference.

\subsubsection{Final Answer Evaluation and Metric Fusion}
\label{app:eval_fusion}

We evaluate final-answer correctness independently from process quality. For problems with textual final answers, Qwen3.5-9B is used as an LLM judge to determine whether the predicted answer is equivalent to the ground-truth answer. For drawing tasks or other tasks whose final answer is visual, we evaluate the generated final image using the same visual matching procedure described above.

For process evaluation, we fuse the local alignment score and the global coverage score using a weighted root mean square:
\begin{equation}
S_{\mathrm{fused}}
=
\sqrt{
\frac{
w_s S_{\mathrm{local}}^2
+
w_j S_{\mathrm{global}}^2
}{
w_s+w_j
}
},
\label{eq:fused-score}
\end{equation}
where $w_s$ and $w_j$ denote the weights assigned to local alignment and global coverage, respectively. By default, we set $w_s=w_j=0.5$.

Finally, many STEM problems admit multiple valid solution paths. Let
$\mathbb{G}=\{\mathcal{G}_1,\ldots,\mathcal{G}_K\}$ denote the set of annotated reference trajectories for a problem. We evaluate the predicted trajectory against each reference trajectory independently and take the best score:
\begin{equation}
S_{\mathrm{final}}
=
\max_{\mathcal{G}_k \in \mathbb{G}}
S_{\mathrm{fused}}(\mathcal{G}_k,\mathcal{P}).
\end{equation}
This best-over-references strategy avoids penalizing a model for following a coherent alternative solution path, while still requiring the predicted trajectory to align with at least one complete and valid reference solution.

\subsubsection{Human Evaluation}
\label{sec:appendix_human_eval_def}

To further validate whether the proposed automatic step-level evaluation is consistent with expert judgement, we conduct a human evaluation on four representative models: Gemini 2.5 Flash Image, GPT-5.4 + GPT-Image, Gemini 3.1 Pro, and Claude Opus 4.6. These models cover both unified image-text generation systems and understanding-only systems. We select 20 {\bench} examples and ask three human experts to independently evaluate the predicted reasoning trajectory of each model.

Given a reference solution containing $N$ ground-truth reasoning steps, each expert identifies the largest prefix length $m$ such that the model prediction correctly covers the first $m$ ground-truth steps in the intended logical order. The normalized human score is then computed as
\[
S_{\mathrm{human}}=\frac{m}{N}.
\]
For samples with multiple valid reference solutions, we follow the same best-match principle as the automatic protocol and use the highest score over the available references.

\section{Implementation Details}
\label{sec:imp_details}

To ensure complete reproducibility of our evaluation pipeline, we provide the detail of the hyperparameters and configuration settings used throughout the experiments.

\subsection{Model Inference Configuration}

All baseline models are evaluated under a strict zero-shot setting. The generation temperature is set to $0.2$ with a top-p of $0.9$, which approximates greedy decoding, and the maximum generation limit is capped at $32\text{k}$ tokens. We apply a unified system prompt enforcing a structured, interleaved output format.

\subsection{Local Evaluator Configuration}

For textual step alignment, we deploy a local Qwen3.5-9B model and aggregate the attention weights. The temperature parameter $\tau$ for attention normalization is set to $0.3$. For image step alignment, we utilize a pre-trained ResNet-50~\cite{DBLP:conf/cvpr/HeZRS16} encoder. The diversity penalty hyperparameter $\lambda$ is set to $0.2$.

\subsection{Global Judge and Metric Fusion}

The LLM-based global judge for assessing whole-trace coverage is also powered by Qwen3.5-9B. To ensure stable and deterministic judgments, we configure it with a temperature of $0.0$, a top-p of $1.0$, and a maximum token length of $512$. In the calculation of the fused process score $S_{\text{fused}}$, the Local and Global scores are assigned equal importance ($w_s = 0.5$ and $w_j = 0.5$).

\section{Baseline MLLMs}
\label{sec:appendix_models}

\subsection{Closed-source Unified Models}

\paragraph{GPT-5.4 + GPT-Image}~\cite{openai_gpt54_2026,openai_gpt_image_1} A pipeline approach that explicitly couples OpenAI's advanced large language model (GPT-5.4) with its image generation counterpart (GPT-Image) to handle complex interleaved text-and-image reasoning tasks.

\paragraph{Gemini 2.5 Flash Image}~\cite{google_gemini25_flash_image_2025} A leading commercial multimodal model from Google designed with native capabilities to seamlessly process and generate both textual and visual outputs within a unified inference process.

\subsection{Open-Source Unified / Interleaved Generation Models}

\paragraph{Janus-Pro}~\cite{janus} An advanced open-source model designed for unified multimodal understanding and generation, operating over both visual and textual modalities in a single framework.

\paragraph{OmniGen2}~\cite{omnigen2} An open-source multimodal model that integrates a diffusion-based image decoder with LLM hidden states, enabling context-driven image generation and visual reasoning without compromising the model's core textual capabilities.

\paragraph{Showo2}~\cite{show} A native unified multimodal model that innovatively fuses autoregressive modeling (for text and understanding tasks) and flow matching/diffusion modeling (for visual generation) within a single Transformer architecture.

\paragraph{Bagel}~\cite{bagel} An open-source foundation model featuring a Mixture-of-Transformer-Experts (MoT) architecture. It dedicates separate expert networks to understanding and generation tasks while sharing a unified self-attention mechanism to efficiently handle interleaved text and image data.

\subsection{Understanding-only Models}

\paragraph{Claude Opus 4.6}~\cite{anthropic_claude_opus46_2026} Anthropic's flagship model known for exceptional logical reasoning and visual understanding capabilities; however, it operates strictly within a vision-text-to-text paradigm.

\paragraph{Gemini 3.1 Pro}~\cite{google_gemini31_pro_2026}  Google's advanced multimodal understanding model. It excels at processing complex multimodal inputs to produce detailed textual reasoning traces but cannot explicitly generate visual evidence.

\paragraph{Qwen Family (Qwen3.5-122B, Qwen3.5-Plus, Qwen3-VL-235B)}~\cite{qwen2026qwen35,qwen3vl} A powerful series of models developed by the Qwen team. While possessing strong foundational language and vision-language understanding, they are fundamentally constrained to text-only generation and rely purely on textual or symbolic reasoning.

\section{Detailed Experimental Results on Each Specific Domain}

Tables~\ref{tab:domain_eng_metrics} --~\ref{tab:domain_bio_oth_metrics} present the detailed results on the Engineering, Physics, Chemistry, Biology, Mathematics, and Other subsets. Overall, the subset-level analysis is consistent with the main findings in the paper: model performance varies substantially across disciplines, and final-answer accuracy alone does not fully reflect multimodal reasoning ability.

A first notable pattern is the consistent advantage of unified models on image-involved reasoning. On Engineering, Physics, and Chemistry, Gemini 2.5 Flash Image and GPT-5.4 + GPT-Image achieve the strongest image-side process scores. This result indicates that tightly coupling visual understanding and generation is beneficial for problems that require intermediate visual derivations or image-based final outputs.
\FloatBarrier
In contrast, understanding-only models remain highly competitive on subsets that favor textual or symbolic reasoning. This trend is particularly clear on Mathematics subset, where models such as Qwen3.5-122B, Qwen3.5-Plus, and Claude Opus 4.6 achieve the best final-answer accuracy despite scoring zero on image-related metrics. These results suggest that when a task can be largely solved through text-based reasoning after extracting key information, stronger language reasoning may outweigh the absence of image generation. At the same time, their consistently zero image-answer accuracy highlights a structural limitation of this model family on tasks whose reasoning process or final output must be expressed visually.

\section{Detailed Experimental Results on Human Evaluation}
\label{sec:appendix_human_eval}

\begin{figure*}[t]
    \centering
    \includegraphics[width=\linewidth]{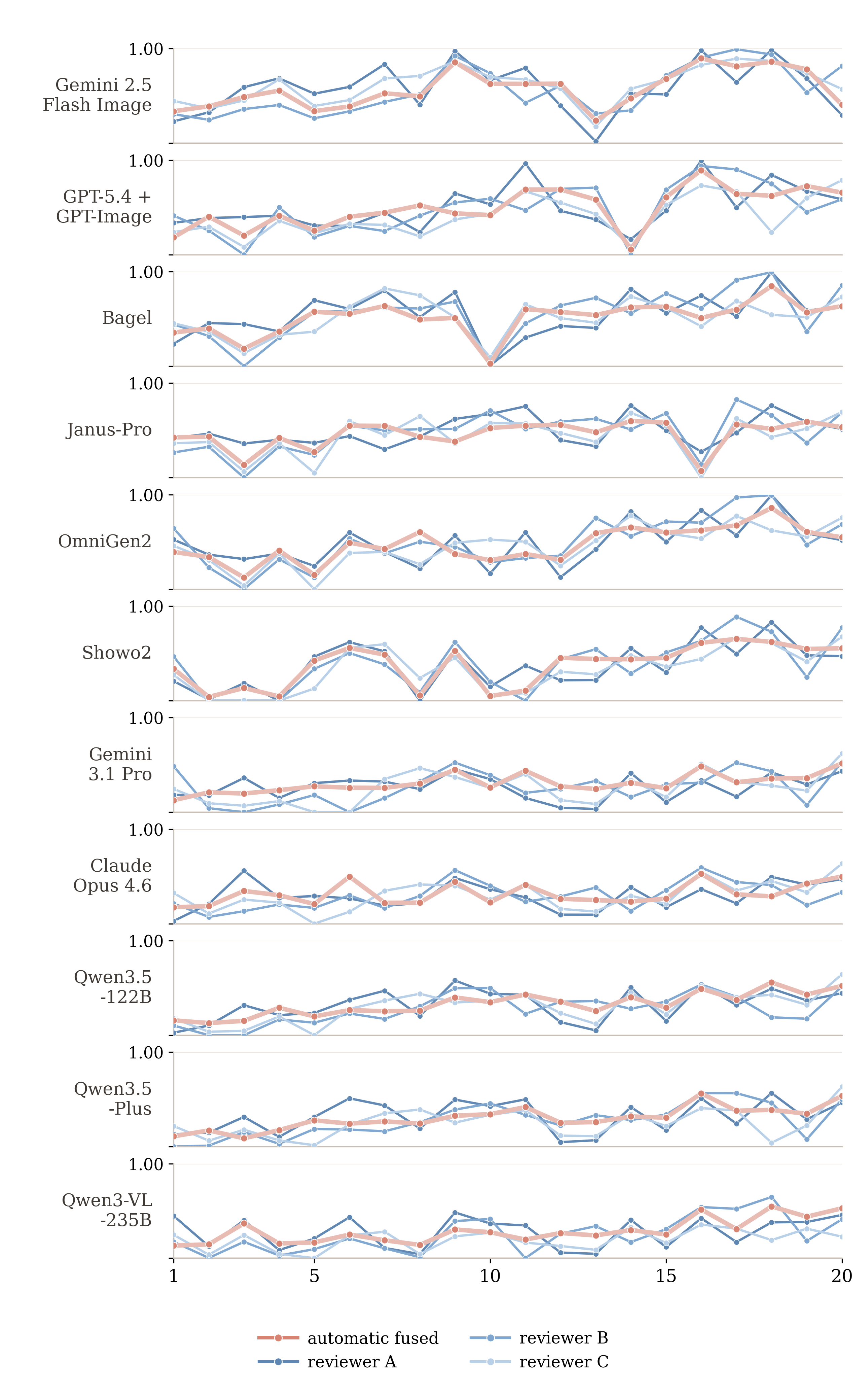}
    \caption{Human evaluation results across all evaluated models. The automatic fused score is compared with the scores assigned by three human reviewers over the same set of examples.}
    \label{fig:human_eval_all_models}
\end{figure*}
Figure~\ref{fig:human_eval_all_models} further presents the human evaluation results across all evaluated models. Overall, the automatic fused score shows strong qualitative agreement with the scores from the three human reviewers. Across different samples, the automatic curve generally follows the rises and drops of human annotations, indicating that our automatic evaluation is broadly consistent with expert judgement. Meanwhile, the three human reviewers exhibit moderate variation, reflecting the inherent subjectivity of judging complex multimodal STEM reasoning traces. Nevertheless, the automatic scores are generally close to the range of expert scores. These results further validate the proposed step-level automatic evaluation as an effective and scalable proxy for expert assessment of multimodal reasoning processes.

\clearpage

\begin{table*}[t]
  \centering
  \small
  \caption{Model performance on the Engineering subset which includes 105 problems with textual final answer and 6 problems with image-based final answer.}
  \label{tab:domain_eng_metrics}
  \resizebox{\textwidth}{!}{
  \begin{tabular}{lccccccccc}
    \toprule
    Model & Acc. & Img. Acc. & Fused & Local-T & Local-I & Local & Global-T & Global-I & Global \\
    \midrule
    \multicolumn{10}{c}{\textit{Unified Models}} \\
    \midrule
    Gemini 2.5 Flash Image & 11.42 & 69.55 & 68.57 & 58.47 & 78.70 & 68.57 & 62.83 & 82.65 & 70.58 \\
    GPT-5.4 + GPT-Image & 33.20 & 55.02 & 69.55 & 57.44 & 59.72 & 57.69 & 73.04 & 89.32 & 79.34 \\
    \midrule
    \multicolumn{10}{c}{\textit{Interleaved Generation Models}} \\
    \midrule
    Bagel & 1.90 & 0.00 & 43.72 & 28.99 & 52.57 & 41.46 & 47.01 & 44.07 & 45.77 \\
    Janus-Pro & 0.00 & 0.00 & 33.37 & 21.09 & 47.30 & 43.70 & 28.51 & 11.24 & 21.32 \\
    OmniGen2 & 0.96 & 0.00 & 43.89 & 26.52 & 60.44 & 41.74 & 45.07 & 38.90 & 45.88 \\
    Showo2 & 0.00 & 0.00 & 39.69 & 37.03 & 56.93 & 48.56 & 34.19 & 24.56 & 30.85 \\
    \midrule
    \multicolumn{10}{c}{\textit{Understanding-only Models}} \\
    \midrule
    Gemini 3.1 Pro & 33.33 & 0.00 & 40.67 & 66.22 & 0.00 & 38.34 & 74.86 & 0.00 & 42.94 \\
    Claude Opus 4.6 & 31.43 & 0.00 & 38.66 & 61.01 & 0.00 & 35.49 & 73.72 & 0.00 & 41.92 \\
    Qwen3.5-122B & 23.81 & 0.00 & 36.06 & 64.12 & 0.00 & 36.81 & 63.33 & 0.00 & 35.22 \\
    Qwen3.5-Plus & 21.90 & 0.00 & 35.35 & 57.10 & 0.00 & 32.91 & 67.22 & 0.00 & 37.68 \\
    Qwen3-VL-235B & 15.96 & 0.00 & 32.84 & 64.05 & 0.00 & 35.57 & 53.97 & 0.00 & 30.16 \\
    \bottomrule
  \end{tabular}
  }
\end{table*}

\begin{table*}[t]
  \centering
  \small
  \caption{Model performance on the Physics subset which includes 93 problems with textual final answer and 1 problem with image-based final answer.}
  \label{tab:domain_phy_metrics}
  \resizebox{\textwidth}{!}{
  \begin{tabular}{lccccccccc}
    \toprule
    Model & Acc. & Img. Acc. & Fused & Local-T & Local-I & Local & Global-T & Global-I & Global \\
    \midrule
    \multicolumn{10}{c}{\textit{Unified Models}} \\
    \midrule
    Gemini 2.5 Flash Image & 9.68 & 0.00 & 70.05 & 61.08 & 66.60 & 64.66 & 70.18 & 80.59 & 75.16 \\
    GPT-5.4 + GPT-Image & 24.63 & 74.18 & 77.36 & 61.58 & 69.62 & 66.74 & 77.98 & 94.96 & 86.45 \\
    \midrule
    \multicolumn{10}{c}{\textit{Interleaved Generation Models}} \\
    \midrule
    Bagel & 3.23 & 0.00 & 48.67 & 30.01 & 52.66 & 43.89 & 53.21 & 52.93 & 53.16 \\
    Janus-Pro & 0.00 & 0.00 & 42.85 & 30.41 & 56.75 & 53.97 & 36.01 & 16.31 & 27.30 \\
    OmniGen2 & 2.15 & 0.00 & 44.66 & 27.57 & 58.76 & 45.90 & 40.80 & 44.15 & 43.51 \\
    Showo2 & 4.30 & 0.00 & 41.95 & 33.67 & 49.93 & 44.73 & 47.48 & 28.19 & 38.82 \\
    \midrule
    \multicolumn{10}{c}{\textit{Understanding-only Models}} \\
    \midrule
    Gemini 3.1 Pro & 33.33 & 0.00 & 37.82 & 68.34 & 0.00 & 37.30 & 70.25 & 0.00 & 38.41 \\
    Claude Opus 4.6 & 35.48 & 0.00 & 38.44 & 63.05 & 0.00 & 34.57 & 77.02 & 0.00 & 41.81 \\
    Qwen3.5-122B & 35.48 & 0.00 & 38.00 & 67.39 & 0.00 & 36.05 & 73.76 & 0.00 & 39.98 \\
    Qwen3.5-Plus & 24.73 & 0.00 & 35.44 & 55.82 & 0.00 & 29.77 & 73.81 & 0.00 & 40.21 \\
    Qwen3-VL-235B & 23.78 & 0.00 & 34.59 & 67.32 & 0.00 & 34.84 & 62.85 & 0.00 & 34.23 \\
    \bottomrule
  \end{tabular}
  }
\end{table*}

\begin{table*}[t]
  \centering
  \small
  \caption{Model performance on the Chemistry subset which includes 23 problems with textual final answer and 2 problems with image-based final answer.}
  \label{tab:domain_che_metrics}
  \resizebox{\textwidth}{!}{
  \begin{tabular}{lccccccccc}
    \toprule
    Model & Acc. & Img. Acc. & Fused & Local-T & Local-I & Local & Global-T & Global-I & Global \\
    \midrule
    \multicolumn{10}{c}{\textit{Unified Models}} \\
    \midrule
    Gemini 2.5 Flash Image & 21.74 & 59.06 & 67.02 & 55.21 & 66.72 & 62.72 & 56.93 & 82.00 & 70.98 \\
    GPT-5.4 + GPT-Image & 30.31 & 73.68 & 68.94 & 53.39 & 55.35 & 56.64 & 65.72 & 86.98 & 78.90 \\
    \midrule
    \multicolumn{10}{c}{\textit{Interleaved Generation Models}} \\
    \midrule
    Bagel & 4.35 & 0.00 & 40.61 & 26.89 & 43.21 & 39.00 & 41.73 & 42.33 & 42.02 \\
    Janus-Pro & 4.35 & 0.00 & 48.97 & 21.06 & 58.41 & 52.02 & 50.00 & 39.67 & 45.82 \\
    OmniGen2 & 8.70 & 0.00 & 44.61 & 19.11 & 58.00 & 43.04 & 39.60 & 53.33 & 46.13 \\
    Showo2 & 0.00 & 0.00 & 43.52 & 32.69 & 55.04 & 46.50 & 49.60 & 26.00 & 40.45 \\
    \midrule
    \multicolumn{10}{c}{\textit{Understanding-only Models}} \\
    \midrule
    Gemini 3.1 Pro & 43.48 & 0.00 & 35.44 & 63.41 & 0.00 & 33.87 & 68.80 & 0.00 & 36.86 \\
    Claude Opus 4.6 & 39.13 & 0.00 & 34.80 & 56.59 & 0.00 & 30.29 & 72.27 & 0.00 & 38.85 \\
    Qwen3.5-122B & 30.43 & 0.00 & 34.95 & 61.70 & 0.00 & 32.74 & 68.73 & 0.00 & 36.91 \\
    Qwen3.5-Plus & 30.43 & 0.00 & 33.28 & 53.48 & 0.00 & 28.66 & 69.80 & 0.00 & 37.43 \\
    Qwen3-VL-235B & 20.40 & 0.00 & 31.79 & 61.63 & 0.00 & 31.64 & 58.57 & 0.00 & 31.60 \\
    \bottomrule
  \end{tabular}
  }
\end{table*}

\begin{table*}[t]
  \centering
  \small
  \caption{Model performance on the Mathematics subset which includes 30 problems with textual final answer.}
  \label{tab:domain_mat_metrics}
  \resizebox{\textwidth}{!}{
  \begin{tabular}{lccccccccc}
    \toprule
    Model & Acc. & Img. Acc. & Fused & Local-T & Local-I & Local & Global-T & Global-I & Global \\
    \midrule
    \multicolumn{10}{c}{\textit{Unified Models}} \\
    \midrule
    Gemini 2.5 Flash Image & 36.67 & 0.00 & 59.33 & 65.61 & 0.00 & 34.83 & 74.83 & 76.94 & 76.21 \\
    GPT-5.4 + GPT-Image & 68.70 & 0.00 & 72.82 & 59.53 & 62.07 & 62.06 & 81.71 & 83.31 & 81.90 \\
    \midrule
    \multicolumn{10}{c}{\textit{Interleaved Generation Models}} \\
    \midrule
    Bagel & 6.67 & 0.00 & 58.01 & 41.54 & 68.40 & 58.12 & 58.28 & 58.89 & 57.99 \\
    Janus-Pro & 0.00 & 0.00 & 49.46 & 25.80 & 56.31 & 53.43 & 64.44 & 24.72 & 45.06 \\
    OmniGen2 & 10.00 & 0.00 & 51.43 & 12.03 & 58.20 & 35.38 & 70.78 & 63.89 & 67.62 \\
    Showo2 & 3.33 & 0.00 & 46.33 & 42.12 & 56.92 & 53.69 & 41.17 & 33.33 & 37.42 \\
    \midrule
    \multicolumn{10}{c}{\textit{Understanding-only Models}} \\
    \midrule
    Gemini 3.1 Pro & 63.33 & 0.00 & 38.74 & 71.34 & 0.00 & 37.55 & 76.61 & 0.00 & 40.02 \\
    Claude Opus 4.6 & 70.00 & 0.00 & 41.51 & 68.51 & 0.00 & 36.23 & 87.44 & 0.00 & 46.13 \\
    Qwen3.5-122B & 76.67 & 0.00 & 40.53 & 70.38 & 0.00 & 37.19 & 83.61 & 0.00 & 43.71 \\
    Qwen3.5-Plus & 70.00 & 0.00 & 37.54 & 64.29 & 0.00 & 34.00 & 77.89 & 0.00 & 40.64 \\
    Qwen3-VL-235B & 51.39 & 0.00 & 36.86 & 70.30 & 0.00 & 35.94 & 71.25 & 0.00 & 37.43 \\
    \bottomrule
  \end{tabular}
  }
\end{table*}

\begin{table*}[t]
  \centering
  \small
  \caption{Model performance on the Other subset which includes 18 problems with textual final answer and 5 problems with image-based final answer.}
  \label{tab:domain_bio_oth_metrics}
  \resizebox{\textwidth}{!}{
  \begin{tabular}{lccccccccc}
    \toprule
    Model & Acc. & Img. Acc. & Fused & Local-T & Local-I & Local & Global-T & Global-I & Global \\
    \midrule
    \multicolumn{10}{c}{\textit{Unified Models}} \\
    \midrule
    Gemini 2.5 Flash Image & 27.78 & 35.16 & 57.60 & 43.87 & 64.18 & 58.37 & 57.25 & 48.91 & 54.16 \\
    GPT-5.4 + GPT-Image & 22.13 & 0.00 & 58.32 & 45.65 & 46.89 & 47.05 & 67.01 & 63.02 & 67.09 \\
    \midrule
    \multicolumn{10}{c}{\textit{Interleaved Generation Models}} \\
    \midrule
    Bagel & 27.78 & 0.00 & 53.26 & 31.04 & 61.14 & 51.43 & 53.62 & 47.47 & 52.74 \\
    Janus-Pro & 0.00 & 0.00 & 50.42 & 15.14 & 57.43 & 51.59 & 53.26 & 39.49 & 47.46 \\
    OmniGen2 & 5.55 & 0.00 & 44.32 & 13.03 & 55.80 & 38.51 & 46.01 & 55.44 & 50.05 \\
    Showo2 & 0.00 & 0.00 & 48.17 & 28.66 & 53.16 & 45.96 & 44.93 & 43.12 & 44.25 \\
    \midrule
    \multicolumn{10}{c}{\textit{Understanding-only Models}} \\
    \midrule
    Gemini 3.1 Pro & 44.45 & 0.00 & 29.28 & 52.87 & 0.00 & 27.68 & 56.52 & 0.00 & 29.84 \\
    Claude Opus 4.6 & 38.89 & 0.00 & 30.37 & 52.27 & 0.00 & 27.44 & 59.42 & 0.00 & 31.99 \\
    Qwen3.5-122B & 33.34 & 0.00 & 29.59 & 52.56 & 0.00 & 27.17 & 55.44 & 0.00 & 30.10 \\
    Qwen3.5-Plus & 22.22 & 0.00 & 30.18 & 50.95 & 0.00 & 26.08 & 63.41 & 0.00 & 33.36 \\
    Qwen3-VL-235B & 22.35 & 0.00 & 25.82 & 52.50 & 0.00 & 26.26 & 47.24 & 0.00 & 25.77 \\
    \bottomrule
  \end{tabular}
  }
\end{table*}

\FloatBarrier
\section{Actual Prompt Templates}
\label{sec:appendix_prompt}

In this section, we provide the prompt templates used in our experiments for reference and reproducibility. These include the prompts for interleaved generation, which instruct models to combine textual and visual reasoning within a unified inference process, as well as the prompts used in the LLM-as-a-Judge evaluation framework. We release these templates to improve the transparency of our experimental setup and to support future research on multimodal reasoning and fine-grained evaluation.

\begin{promptbox}{1. INTERLEAVED\_QUERY\_PREFIX}
You will be given a problem. Please answer by strictly following these instructions:

[Task Requirements]
1. Carefully understand the problem first, then provide a complete step-by-step solution.
2. The solution process must use an "interleaved text-and-visual" format:
- For each key step, first explain the reasoning in text.
- Then provide a "visual illustration" block to support that step.
- Each "Visual Illustration" must be an actual image output generated by the model. It must not be replaced by ASCII art, tables, plain text diagrams, flowcharts written in text, or any other text-only substitute.
3. At the end, provide the final answer separately, and it must be strictly wrapped in:
<final_answer>final answer</final_answer>
4. Do not put any extra explanation inside the `<final_answer>` tag other than the final answer itself.
5. Any requirement in the problem statement about the form of the final submitted answer applies only to the final answer itself, not to the reasoning process before it.
6. Any answer-format requirement in the problem statement constrains only the content inside `<final_answer>...</final_answer>'. It does not constrain the length, level of detail, or structure of the reasoning before it.

[Required Output Format]
- Use the following structure:

Problem Understanding
Textual Reasoning
Visual Illustration
Textual Reasoning
Visual Illustration
...
Final Answer
<final_answer>...</final_answer>

[Style Requirements]
- Start directly with "Problem Understanding"; do not describe user intent or restate instructions.
- Do not output meta narration (e.g., "The user wants...", "I need to...", "Let's...").
- Use only the required section headers above. Do not add extra sections, preface, appendix, or commentary.
- Do not output any headings like "Step 1", "Step 2", etc.
- Fully expand the key reasoning needed to justify the answer. Do not over-compress intermediate derivations.
- Prefer preserving explicit intermediate steps over aggressive summarization.
- Each "Textual Reasoning" block may use multiple sentences when needed to make the reasoning complete and self-contained.
- Each "Visual Illustration" block must be an actual image, not a text block.
- Each required section header should appear on its own line.
- Each "Textual Reasoning" block should be followed by a line break before the next "Visual Illustration" block.
- If the solution naturally contains multiple key substeps, keep them as separate interleaved blocks instead of merging them into one paragraph.
- Each visual illustration must closely match the corresponding step.
- Do not omit key reasoning or derivation steps.

[Hard Constraint]
- <final_answer>...</final_answer> must appear exactly once and be the last non-empty line.
- The last non-empty line must be exactly one block: <final_answer>...</final_answer>.
- Do not output any text after the final answer tag.
- Do not use ASCII art or text-only pseudo-visuals in place of images.
- If information is insufficient, still output <final_answer>insufficient_information</final_answer>.
\end{promptbox}

\begin{promptbox}{2. TOOL\_INTERLEAVED\_PROMPT}
You are solving a multimodal problem.

Please solve it through a genuinely interleaved text-and-image reasoning process. Your goal is not to finish as quickly as possible, but to use generated images as part of the reasoning itself.

Follow these principles:

1. Produce at least one text reasoning segment.
2. Generate multiple genuinely helpful images and distribute them throughout the reasoning process.
3. Text and images must be interleaved:
   - do not write a long stretch of text first and then dump several images at the end;
   - do not generate multiple images in a row without new reasoning progress between them.
4. The process should look like:
   reasoning -> helpful image -> further reasoning from that image -> another helpful image -> further reasoning ...
5. Every image must directly support reasoning rather than decoration. It may help with:
   - understanding the setup
   - marking key objects
   - highlighting relevant regions
   - clarifying structural relations
   - redrawing a schematic more clearly
   - showing an intermediate analytical result
6. Each image must advance the later reasoning, rather than merely repeat earlier content or serve as a presentation illustration.
7. Before generating an image, explain what is still uncertain, what the image is meant to clarify, and why it is useful.
8. After an image is generated, read it carefully and continue reasoning from what it clarifies.
9. The reasoning must be sufficiently detailed and clear. Do not give only a one-line conclusion.
10. Do not generate images that merely paste large amounts of text on the canvas. Images must provide genuinely useful visual information such as schematics, structural relations, arrows, axes, regions, force directions, connections, or key labels.
11. Any answer-format requirement in the problem statement constrains only the final answer itself. It does not constrain the length, level of detail, or structure of the reasoning before it.
12. Requirements such as "one token", "ASCII only", "no extra text", "one final image only", "exact format", or similar restrictions apply only to the final answer, not to the earlier reasoning process.
13. A bare short answer string by itself is not a valid reasoning process. Before the final answer, provide at least one substantive reasoning block even for easy problems.
14. If the problem includes an image, do not end immediately with only a short final answer unless the reasoning has already been explained in a substantive text block.
15. Only give the final answer when you believe the conclusion is stable.
16. For text-answer problems, the final answer must appear in the final text portion and must end with exactly one <final_answer>...</final_answer> block.
17. Do not output any text after the final answer tag.
18. If the required final answer is itself an image, structure, diagram, conformation, or drawing, then the last generated image should be the answer image itself rather than merely an auxiliary illustration.
19. Even when the final answer is an image, provide at least one substantive text block before the answer image explaining what is being drawn, what key structure or relation it must show, and why it is the correct final answer.
20. For such drawing problems, do not replace the answer image with markdown image syntax, SVG code, a data URI, an attachment placeholder, or any other text surrogate. The final answer should be realized as the last generated image itself.
21. Unless the problem is truly trivial, do not finish before using multiple helpful generated images to advance the reasoning.
\end{promptbox}

\begin{promptbox}{3. Final-answer Judge Prompts}
You are a strict and careful judge for final-answer equivalence. Your only job is to decide whether the predicted final answer is truly semantically identical to one acceptable ground-truth answer. Be conservative: when a difference may change the meaning, treat it as not equivalent.

----------------------------------------

You are judging final-answer equivalence.

Task: Decide whether the predicted final answer should count as correct against at least one acceptable ground-truth answer.

Decision policy: Be conservative. Mark the answer correct only when the predicted final answer and a ground-truth answer have the same meaning with no material mismatch. If you are unsure, return 0.

Important rules:
1. Focus on final-answer equivalence only.
2. Ignore harmless formatting differences only: punctuation, whitespace, LaTeX wrappers, capitalization, surrounding filler words, and standard algebraic rewrites.
3. Give credit for mathematically equivalent expressions only when they are fully equivalent, with the same variables, same constraints, same set of solutions, same sign, same units, and same boundaries.
4. Do not give credit when any meaningful content changes, including:
   - different numeric value or approximation that is not clearly intended as the same value
   - different sign, inequality direction, interval boundary, ordering, or logical relation
   - different variable, symbol, label, object identity, or named entity
   - different chemical structure, substituent, isotope placement, compound name, or mapping between labels such as A/B/C/D/E
   - different biological sequence, gene/protein name, stage, category, or ordering
   - one answer being more vague, more generic, or only partially matching the ground truth
5. For structured answers with multiple items, all items must align correctly. A reordering is acceptable only if the answer is explicitly unordered and every item still matches exactly. If labels or item-to-item correspondences change, return 0.
6. Do not infer correctness from topic similarity. Two answers in the same domain are still wrong if any key term or component differs.
7. Use the candidate response tail only as supporting context when the extracted predicted final answer is awkwardly formatted. Do not use it to invent a better final answer than the extracted one.
8. If the predicted answer is empty, image-only, placeholder-like, or lacks enough information to verify exact equivalence, return 0.
9. Output exactly one block between <judge_result> and </judge_result>.
10. Inside the block output a single JSON object with fields:
   "verdict": 0 or 1
   "matched_gt_index": integer index starting from 1, or 0 if none
   "reason": short string

Checklist before returning 1:
- Same objects or entities?
- Same mapping between labels and items?
- Same mathematical meaning, not just similar form?
- No missing or extra components?
If any answer is "no" or "unclear", return 0.

Problem: {question_text}
Predicted extracted final answer: {pred_block}
Candidate response tail: {raw_block}
Acceptable ground-truth final answers: {gt_block}
\end{promptbox}

\begin{promptbox}{4. Whole-trace Judge Prompts}
You are a generous, evidence-based judge for reasoning-process coverage.

----------------------------------------

Task: Given the original problem, a candidate model's whole reply, and a list of reference contents, determine for each reference content whether the candidate reply semantically covers it.

Core principle: Your goal is to judge semantic entailment and coverage, not literal wording overlap. Prefer avoiding false negatives on actually correct reasoning. When the candidate reply is mathematically consistent and clearly on the right solution path, be willing to count semantically implied references as covered. Treat a clearly correct final derivation as positive evidence for intermediate coverage whenever the missing intermediate item is a standard consequence of that derivation.

Judging rules:
1. Match against the entire candidate reply, not only the final answer.
2. Mark 1 if the candidate reply explicitly states the reference content, clearly paraphrases it, or provides enough mathematical or semantic evidence that entails it.
3. Mark 1 if the candidate reply gives an algebraically equivalent formula, an equivalent constraint, an equivalent domain statement, or a stronger statement that clearly subsumes the reference.
4. Mark 1 if the evidence is distributed across multiple nearby or logically connected sentences or equations, as long as together they support the reference content.
5. Mark 1 if a later derivation, equation, or final expression clearly implies the reference content, even if the intermediate wording is omitted.
6. Mark 1 if the candidate reply reaches the same local conclusion or a directly equivalent stronger conclusion by a different but valid derivation path.
7. Do not require the same wording, the same notation, the same variable names, the same derivation order, or the same decomposition granularity as the reference.
8. Ignore harmless differences in formatting, equation rearrangement, symbol renaming, simplification, factorization, unit style, and equivalent notation.
9. If the candidate reply states a more concrete result that logically contains the reference idea, count the reference as covered.
10. Use a generous semantic matching standard for correct or near-correct reasoning traces; do not demand verbatim intermediate steps.
11. If the candidate reply provides the correct equation, invariant, constraint, ranking, or final symbolic result from which the reference naturally follows, count it as covered even when the explicit wording differs.
12. Mark 0 only if the reference content is truly missing, too vague to verify, merely topically related, or contradicted by the candidate reply.
13. Do not give 1 for generic topic overlap without concrete supporting evidence.
14. When uncertain between 0 and 1, prefer 1 if there is plausible mathematical or semantic evidence in the candidate reply.

Output format:
- Output exactly one block between <judge_result> and </judge_result>.
- Inside the block, output one line per reference in the format step_id=0 or step_id=1.
- Do not output explanations.

Problem: {question_text}
Candidate whole reply: {raw_response}
Reference contents: {ref_block}

----------------------------------------

You are a careful multimodal judge for reasoning-process coverage.

Task: Given the original problem, a candidate model's whole reply, and a GT reference image step, determine whether the candidate reply semantically covers the visual reasoning evidence represented by the reference step.

You will receive multiple images as native model inputs in this exact order: {image_inputs_block}

The important regions are specified by normalized bounding boxes over the GT reference image only. You should compare the GT reference image against the candidate generated images and decide whether the candidate generated images support the same visual reasoning evidence or conclusion associated with this GT step.

Judging rules:
1. Consider both the reference text and the important GT image regions.
2. Mark 1 only if at least one candidate generated image semantically matches the GT visual evidence.
3. Pure text alone is not sufficient for image coverage. If the candidate has no relevant generated image evidence, mark 0.
4. Mark 1 if the candidate image evidence plus the candidate reply together capture the same visual idea, relation, geometry, highlight, or conclusion as the GT reference.
5. Mark 0 if the relevant visual evidence is missing from the candidate image panels, contradicted, or too vague to verify.
6. The question image, if provided, is context only. Do not count it as candidate evidence.
7. When uncertain between 0 and 1, prefer 0 unless there is visible evidence in the candidate image panels.

Output format:
- Output exactly one block between <judge_result> and </judge_result>.
- Inside the block, output exactly one line in the format {step_id}=0 or {step_id}=1.
- Do not output explanations.

Problem: {question_text}
Candidate whole reply: {raw_response}
Reference image step id: {step_id}
Reference text: {reference_text}
Important GT image regions: {bbox_block}
\end{promptbox}

\begin{promptbox}{5. Score Text-Matching Prompt}
You are a strict grader.

Your goal is to determine whether the Predicted reasoning step between <predicted> and </predicted> semantically contains the Reference key point between <reference> and </reference>.

Judging procedure:
1. Identify the essential propositions in the Reference key point.
2. Check whether each essential proposition is present in the Predicted reasoning step, allowing paraphrase.
3. If all essential propositions are covered and there is no contradiction, output 1.
4. Otherwise output 0.

Rules:
- Paraphrase counts as match.
- Missing any essential proposition => 0.
- Contradiction => 0.
- Mere topic similarity => 0.
- Extra correct information is acceptable.
- Exactly same => 1.
- Same information => 1.

Predicted reasoning step:
<predicted>
\end{promptbox}

\end{document}